# A novel approach for nose tip detection using smoothing by weighted median filtering applied to 3D face images in variant poses

Parama Bagchi
Department of Computer Science & Engineering
MCKV Institute of Engineering
Kolkata 711204, India
paramabagchi@gmail.com

Debotosh Bhattacharjee Department of
Computer Science & Engineering Jadavpur
University
Kolkata-700032, India
debotoshb@hotmail.com

Mita Nasipuri
Department of Computer Science & Engineering
Jadavpur University
Kolkata-700032, India
mitanasipuri@gmail.com

Dipak Kumar Basu
Department of Computer Science & Engineering
Jadavpur University
Kolkata-700032, India
dipakkbasu@gmail.com

*Abstract*— **This paper is based on an application of smoothing of 3D face images followed by feature detection i.e. detecting the nose tip. The present method uses a weighted mesh median filtering technique for smoothing. In this present smoothing technique we have built the neighborhood surrounding a particular point in 3D face and replaced that with the weighted value of the surrounding points in 3D face image. After applying the smoothing technique to the 3D face images our experimental results show that we have obtained considerable improvement as compared to the algorithm without smoothing. We have used here the maximum intensity algorithm for detecting the nose-tip and this method correctly detects the nose-tip in case of any pose i.e. along X, Y, and Z axes. The present technique gave us worked successfully on 535 out of 542 3D face images as compared to the method without smoothing which worked only on 521 3D face images out of 542 face images. Thus we have obtained a 98.70% performance rate over 96.12% performance rate of the algorithm without smoothing. All the experiments have been performed on the FRAV3D database.**

*Keywords- Weighted median filtering,3D face images,noisy spikes.*

## I. INTRODUCTION

In many of the image processing applications including 3d face 3D face recognition, meshes deliver a simple and flexible way to represent and handle complex geometric objects like 3D face mesh grids. Dense triangles meshes are the standard output of modern shape acquisition techniques such as laser scanners. The surface of a 3D model reconstructed from real-world data is often corrupted by noise. An important problem is to suppress noise while preserving the geometric features of the model. In this paper we introduce weighted median filtering scheme for smoothing noisy 3D faces. The main idea of our approach consists of applying weighted median filtering to points on mesh grids and then applying a feature detection algorithm i.e. locating the nose tip on a 3d face in any orientation. In image processing normally, median filters are simple and very effective tools for noise suppressing. The basic idea of the median filtering consists of simultaneous replacing every pixel of an image with the median of the pixels contained in a window around the pixel. Mean filtering is usually used for suppressing Gaussian noise while median filtering is a powerful tool for removing impulsive noise [1]. A new approach has been developed recently namely the iterative mean and median filtering schemes and their modifications became very popular because of their close connection with PDE methods in image processing [10].

In our method from the training set of 3D mesh image in any pose the face images have been smoothed by weighted median filtering and the nose tip has been correctly localized as having the highest intensity value. In Section 2, some related works have been reviewed on 3D face recognition. A facial feature extractor, which uses the proposed smoothing [6] technique, has been introduced in Section 3. Experimental results have been reported in Section 4.A comparative analysis is enlisted in Section 5.Finally conclusions are enlisted in Section 6.

## II. RELATED WORKS

A number of methods have been proposed for denoising and smoothing triangle meshes to produce high quality models. This section presents an overview of these methods for filtering of 3D triangle meshes[3] and an overview of methods for feature detection. One of the first signal processing approaches in the field of triangle meshes filtering was a surface signal low-pass filter design [Taubin 1995] used to modify the vertices positions in a linear complexity algorithm. A mean and a median filter were applied to the normal vector of triangles [Yagou et al. 2002]. Here in this method the normal vector of a triangle was modified according to the normal vectors of its neighbours and then the vertices positions are updated. Based on the same strategy, an adaptive Gaussian filter [Ohtake et al. 2002] and then an adaptive minimum mean squared error filter [Mashiko et al. 2004] instead of the as compared to the algorithm without smoothing filters have been applied to the normal vector of triangles. Yet another adaptive repeated local averaging filter with histogram based filtering procedures [Yagou et al. 2003] has been proposed. An integration method combined with a scalable Laplacian and a curvature flow operator [Desbrun et al.1999] was introduced to improve the diffusion process.This method allows the combination of different types of filters. It also avoids the shrinkage problem usually encountered when using the Laplacian operator.A method based on prescribed mean curvature flow adapted to the preservation of some non-linear geometric features was also introduced. Several of these overviewed methods were inspired by signal processing theory. They work locally and iteratively on the data structure of the surface to modify the vertices positions. They are based on different kinds of low-pass filter designs and some of them [Mashiko et al. 2004; Ohtake et al.2002; Yagou et al. 2003] are adaptive according to one or more local surface parameters to preserve geometric features of the surface. The remaining methods [Bobenko and Schröder 2005; Desbrun et al. 1999; Hildebrandt and Polthier 2004] evaluate higher level information of the surface such as curvature or energy parameters and the filtering algorithms[6] are based on this higher level information. Also a number of significant works have been done on 3D face localization. The paper on 3D face localization by Przemyslaw Szeptycki, Mohsen Ardabilian, and Liming Chen [11] dealt only with curvature analysis in the frontal pose. The paper on curvature analysis published by Przemyslaw Szeptycki, Mohsen Ardabilian, and Liming Chen was feature localization on non- frontal 3D face using curvature analysis. Also 3D face detection using curvature analysis [12], dealt with feature localization using curvature analysis followed by face registration. Although curvature analysis method is a widely adopted technique and is robust to translation and rotation but the problem in majority of cases is that it is prone to noise. Before applying the methods additional filters have to be applied. Another problem is that after applying the method there are multiple points which have maximum and minimum curvatures on the basis of HK (H-mean curvature, K-Gaussian curvature classification).The paper by Fatimah Khalid, Lili N. A.□s[13] tend to identify nose tip on the basis of its□ maximum depth. But in different poses nose tip do not have the highest depth. The paper by D.O. Gorodnichy, S. Gurbuz, K. Kinoshita and S. Kawato[14][15] was on the shading technique and was applied for identifying the nose tip and nose ridge.The limitation of this method was that the illumination along the nose issensitive to the lighting condition. Secondly, the existing method was proposed under the assumption that the nose tip is the highest point in 3D facial data [16]. The work mentioned in[17] used chin neck juncture to be selected manually in advance. Here input data captured by VIVID700 is not sensitive for the black color like hair, because it uses a laser and so the problem was solved by covering the hair with the cap. The paper mentioned [18] had the complexity to compensate spin images. In contrast to the techniques mentioned above, the technique presented here removes all the shortcomings discussed here. Henceforth the problems caused by curvature analysis are not present in our method. At first thresholding has been applied and for that purpose Otsu□s method [19] is employed. Otsu□s method is primarily applied because the problem with global thresholding is that changes in illumination across the scene may cause some parts to be brighter (in the light) and some parts darker (in shadow) in ways that have nothing to do with the objects in the image.Otsu□s method sets the threshold so as to try to make each cluster as tight as possible, thus minimizing their overlap.As we adjust the threshold one way, we increase the spread of one and decrease the spread of the other. The goal then is to select the threshold that minimizes the combined spread. We have defined the within-class variance as the weighted sum of the variances of each cluster. Luckily the threshold with the maximum between class variance also has the minimum within class variance. Primarily thresholding has been applied in our technique so that only the face is used in recognition. The ears, hair and irrelevant details are not taken into consideration. The best part of the present technique is that it is robust to black colors of facial hairs like moustache, facial scars etc. Also the present technique even works fabulously under various illumination conditions. The present technique is pose invariant and correctly localizes the nose tip across different poses. Here we extend the use and propose a generalization of the 2D weighted median filtering technique and we apply the idea to a 3D face image. We have extended the idea of weighted median filtering and used it to smooth the surface of a 3D face and localized the nose tip. So basically we have implemented a 3D weighted median filtering technique based on the local neighbourhood[7]. The problem with the method without smoothing was that since we did not use any smoothing technique, so in case of some 3D face images noisy spikes contributed to points of high intensity. After applying the present technique we obtained formidably good results as is discussed below in Section 4.

III. PROPOSED WORK

A range image (Figure 1) is a set of points in 3D each containing the intensity of it□s pixels. The technique also holds in case of 2.5D images which may be described as containing at least one depth value for every (x, y)coordinate. The acquisition process is described as follows: - Normally, a 3D mesh image is captured by a 3D camera such as a Minolta Vivid 700 camera and a range image is generated from the 3D mesh image. The image in our case is generally in the form of $z = f(x, y)$.Next some pre- processing methods was applied to eliminate unwanted details such as facial hairs, scars etc.

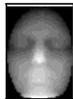

Figure 1. 2.5D range image

After the necessary features are located, the alignment of models is done using some translation and rotation process. In the last and final step a classifier is designed to test the validity of the designed dataset. Normally a procedure for face recognition using range images is composed of four steps(Figure2).

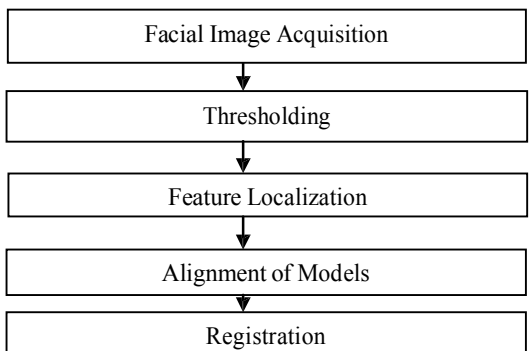

Figure 2. An overview of the system without smoothing

After applying the 3D weighted median filtering an overview of our present system looks as follows:-

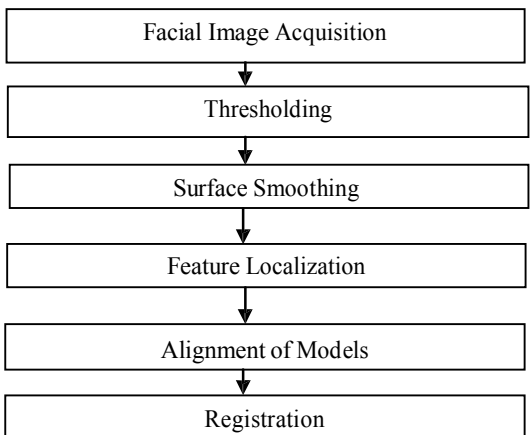

Figure 3.An overview of the system after smoothing

The present technique makes use of the steps below:-
- Facial Image Acquisition
- Thresholding
- Surface Smoothing
- Feature Localization
- Alignment of models

*A. Facial Image Acquisition*

The present technique uses 3D face database FRAV3D database[8]. We have 542 faces consisting of different poses (including rotation about x-axis, y-axis and z-axis). A range image is an array of numbers where the numbers quantify the distances from the focal plane of the sensor to the surfaces of objects within the field of view along rays emanating from a regularly spaced grid. For example, a nose tip is the closest point to the camera on a face, so it has the highest numerical value. Different from 3D mesh images, it is easy to utilize the 3D information of range images because the 3D information of each point is explicit on a regularly spaced grid. Due to these advantages, range images are very promising in face recognition

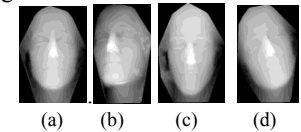

(a)        (b)        (c)        (d)

Figure 4. Samples from the FRAV3D Database(a,b,c,d), where the depth maps corresponding to a single person are shown for frontal pose, image rotated about Y axis, image rotated about X-axis, and images rotated about Z axis.

*B. Thresholding:-*

Thresholding [2] is an important technique for image that tries to identify and extract a target from its background on the basis of the distribution of gray levels or texture in image objects. The simplest method is a gray level image and it☐s background. Pixels with similar value in the neighborhood usually belong to the same region. For a gray level image f(x, y), bi-level thresholding is to transform f(x, y) to binary image g(x, y) by a threshold which can be expressed as:-

$$g(x,y) = 0 \text{ if } f(x,y) < T$$
$$= 1 \text{ if } f(x,y) >= T \ldots\ldots\ldots \quad (1)$$

Otsu☐s method is an adaptive thresholding technique that is applied here. Otsu's thresholding method involves iterating through all the possible threshold values and calculating a measure of spread for the pixel levels each side of the threshold, i.e. the pixels that either falls in foreground or background. Otsu☐s method selects the threshold by minimizing the within-class variance of the two groups of pixels separated by the thresholding operator. After applying Otsu☐s method 3D thresholded mesh grids corresponding to depth maps in Figure4(a),4(b),4(c) and 4(d) are obtained as in Figure5(a),5(b),5(c) and 5(d).

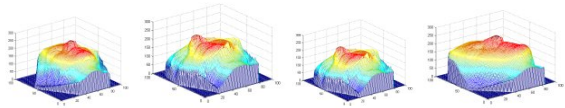

(a) Frontal pose (b) Rotated about y-axis    c) Rotated about x-axis
d) Rotated about z axis
Figure 5. Mesh-grids after thresholding

*C. Surface Smoothing*

Surface smoothing refers to the fact that noisy spikes and other various deviations are sometimes caused on the 3D face image by noise and other several other factors.So some type of smoothing techniques [9] are to be applied. In our present technique we have extended the concept of 2D weighted median filtering technique to 3D face images. The present technique performs filtering of 3D dataset using the weighted median implementation of the mesh median filtering. The weighted median filter is a modification of the simple median filter.

Weighted median filtering: - Weighted Median (WM) filters are the filters that have the robustness and edge preserving capability of the classical median filter and resemble linear FIR filters. In addition weighted median filters belong to the broad class of nonlinear filters called stack filters. This enables the use of the tools developed for the latter class in characterizing and analyzing the behaviour of Weighted Median filters in noise attenuation capability. Figure 6. shows an oriented triangle mesh.

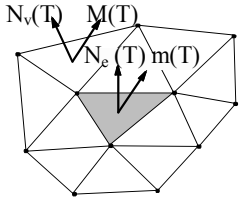

Figure 6. Original mesh triangle

Let T be a mesh triangle, n(T) be the unit normal of T, A(T) be the area of T, and C(T) be the centroid of T of the mesh triangle. Let us divide the set of neighboring triangles of a given triangle in two subsets: -The set of mesh triangles Ne(T), the set of mesh triangles sharing an edge with T and the set of mesh triangles Nv(T) sharing a vertex with T. We have assigned weights 1 to triangles of Ne(T), and weights 1 to triangles of Nv(T) (Figure.7).The triangle in Figure.7 shaded in black is the median. Positive weights tends to smooth the 3D images. On the other hand negative weights distort the noise elements. After the entire filtering operation, we select the median of the mesh triangle

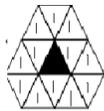

Figure 7. Various weights allocated to triangles of mesh

Figure 8 shows the effect of smoothing.

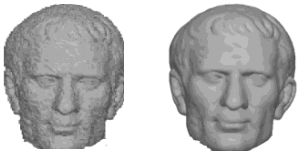

Figure 8.(a) Noisy Model (b)Smoothed Model

The weighted median filtering scheme [1] described below is a simple and useful modification of the basic median filter. Consider a set of samples($x_0$, $x_1$......... $x_{n-1}$) and positive weights ($w_0$, $w_1$,........$w_{n-1}$).It is evident that elements with higher weights are more frequently selected by the median filter. The present algorithm for smoothing works as follows:-

Function Median_3D (Original Mesh grid)
Step1:-Input the face mesh grid.
Step 2:- Initialise a weight matrix w consisting of positive weights with 27 elements
Step 3:- Build the neighbourhood. The neighbourhood of the mesh grid should be a cubic power of an integer (e.g. 3x3x3=27, 5x5x5=125) depending on the filter□s window dimension (3x3x3,5x5x5 and so on) The structuring element coefficients h(i,j,k), i=1,…,N , j =…N and k=1,…,N are written in a row-wise manner. The coordinates i, j, k correspond to 3D coordinates x, y, z. In the present algorithm we have taken the neighbourhood to be consisting of 27 elements.
Step 4:-Perform the processing on the median pixel surroundedby the neighbourhood points.
Step 5:- Now sort the 27 elements
Step6:- Return the median element i.e.the 14th element.
End Function

The algorithm was run for 100 itherations and considerable smoothing was obtained. After smoothing the results obtained are as follows:-

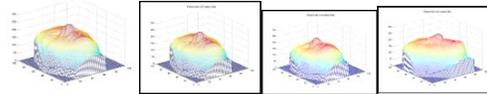

Figure 9. Mesh grids after smoothing

*D. FeatureLocalization*

Faces have a common general form with prominent local structures such as eyes, nose, mouth, chin etc. Facial feature localization is one of the most important tasks of any facial classification system. To achieve fast and efficient classification, it is needed to identify features which are mostly needed for classification task.

• *Surface Generation:* - The next part of the present technique concentrates on generating the surface [13] of this 3D mesh image. For the nose tip localization we have used the maximum intensity concept as the tool for the selection process. Each of the 542 faces (including rotation in any direction in 3D space namely about x-axis, y-axis and z-axis) in the FRAV3D database has been next inspected for localizing the nose tip. A set of fiducial points are extracted from both frontal and various poses of face images using a maximum intensity tracing algorithm. As shown in Figure.10, the nose tips have been labeled on the facial surface, and accordingly, the local regions are constructed based on these points. The maximum intensity algorithm used for our purpose is given :-

Function Find_Maximum_Intensity (Image)
Step 1:-Set max to 0
Step 2:- Run loop for I from 1 to width (Image)
Step 3:- Run loop for J from 1 to height (Image)
Step 4:- Set val to sum(image(I-1:I+1,J-1:J+1))
Step 5:- Check if val is greater than max.
Step 6:- Set val to val2
Step 7: - End if
Step 8:- End loop for I
Step 9:- End loop for J

Then after generation of the 3D surface the results which are obtained are shown:–.

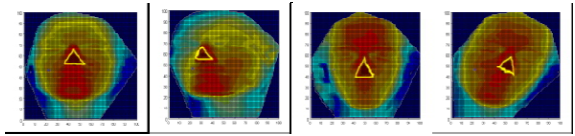

Figureure 10. M. Surface Generation with nose localized

*E . Alignment of Models*:- After feature localization based on the extracted feature we have to align the extracted face model. This alignment [14] with the major axes greatly simplifies the task for registration. These features are then used for coarse alignment and scale normalization. To eliminate the tilt along Y-axis, the image has to be rotated can be obtained by multiplying the original pointcloud image by the matrix given as follows:-

$S_i'' = M_y * S_i$ where

$M_y = \begin{pmatrix} -\sin(\theta) & 0 & \cos(\theta) & 0 \\ 0 & 1 & 0 & 0 \\ -\sin(\theta) & 0 & \cos(\theta) & 0 \\ 0 & 0 & 0 & 1 \end{pmatrix}$

To eliminate the tilt along X-axis, the image has to be rotated can be obtained y multiplying the original pointcloud image by the matrix given as follows:-

$S_i'' = M_x * S_i$ where

$M_x = \begin{pmatrix} 1 & 0 & 0 & 0 \\ 0 & \cos(\theta) & -\sin(\theta) & 0 \\ 0 & \sin(\theta) & \cos(\theta) & 0 \\ 0 & 0 & 0 & 1 \end{pmatrix}$

To eliminate the tilt along Z-axis, the image has to be rotated can be obtained by multiplying the original pointcloud image by the matrix given as follows:-

$S_i'' = M_z * S_i$ where

$M_z = \begin{pmatrix} \cos(\theta) & -\sin(\theta) & 0 & 0 \\ \sin(\theta) & \cos(\theta) & 0 & 0 \\ 0 & 0 & 1 & 0 \\ 0 & 0 & 0 & 1 \end{pmatrix}$

By calculating this matrix for a variety of rotations an orientation which maximizes the symmetry can be chosen. So this pre-processing is actually essential to finding the correspondence between 2 surfaces which would finally help in registration.

## IV. EXPERIMENTAL RESULTS AND DISCUSSION

The testing phase has been performed on the FRAV3D database itself and it contains a comparison of the present algorithm using smoothing as compared to the algorithm without t smoothing. The data set contains 542 faces out of which 282 faces were in neutral pose, 94 faces were rotated about z axis, 94 faces about y axis and 72 faces about x axis. In the following section the first table would show the result on a large set of databases after applying or as compared to the algorithm without smoothing and the second table would show the result after applying smoothing on the set of faces for each section.

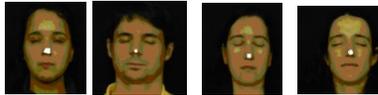

Figure.11. Some samples in frontal pose from FRAV3D Database

*A. Frontal pose:* - The first test was made on the face models in frontal pose only (Figure. 11) (models simulate un-supervised conditions during acquisition). During this test all face models (only frontal poses selected consisting of 282 faces) were accepted and nose tip were correctly localized.

1.Before smoothing:- The images were not smoothed and nose tips correctly recognized.

2.After smoothing: - The images smoothed and nose tips correctly recognized.

| Table 1 | Nose Localization in Frontal Pose | | |
|---|---|---|---|
| | *No of nose tips localized* | *% of Success* | *% of Failures* |
| 1. | 282 | 100% | 0% |

*B. Rotation about Y-Axis:-*

1. Non-frontal pose: - The second test was performed on the face models taking the faces rotated about y axis. (Figure 12)During this test 94 faces (only non-frontal poses rotatedabout y axis were selected) were accepted and nose tip were correctly localized for 85 faces.

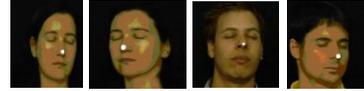

Figure.12. Some samples rotated about Y axis from FRAV3D Database

Before smoothing:-The results obtained before smoothing are shown in the following:-

| Table 2 | Nose Localization about Y axis | | | | |
|---|---|---|---|---|---|
| | *Viewpoint around Y axis* | *No. of nose-tips in each angle* | *No. of nose tips correctly localized* | *% of success* | *%of failures* |
| 1. | +30 | 10 | 8 | 90% | 10% |
| 2 | +30 | 10 | 10 | | |
| 3. | -30 | 21 | 18 | | |
| 4. | +38 | 21 | 21 | | |
| 5. | -38 | 16 | 12 | | |
| 6. | +40 | 16 | 16 | | |

After smoothing:-The results obtained after smoothing are shown

| Table 3 | Nose Localization about Y axis | | | | |
|---|---|---|---|---|---|
| | *Viewpoint around Y axis* | *No. of nose-tips in each angle* | *No. of nose-* | *% of success* | *%of failures* |
| 1 | +30 | 10 | 9 | 96% | 4% |
| 2 | +30 | 10 | 10 | | |
| 3. | -30 | 21 | 20 | | |
| 4. | +38 | 21 | 21 | | |
| 5. | -38 | 16 | 15 | | |
| 6. | +40 | 16 | 16 | | |

Thus we have obtained 96% improvement over the algorithm

*C. Rotation about Z-Axis:-*

1. Non-frontal pose: - The third test was performed on the 94 taking the faces rotated about z axis (Figure 13). During this test all face models (only non-frontal poses rotated about z axis were selected) were accepted and nose tip were correctly localized).

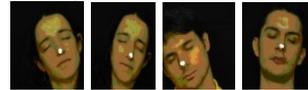

Figure.13. Some samples rotated about z axis from FRAV3D Database

Before smoothing:- The results obtained before smoothing are shown in the following table:-

| Table 4 | Nose Localization about Z axis | | | | |
|---|---|---|---|---|---|
| | *Viewpoint around Z axis* | *No. of nose-tips in each angle* | *No. of nose-* | *% of success* | *%of failures* |
| 1. | +18 | 13 | 13 | 97% | 3% |
| 2. | -18 | 13 | 13 | | |
| 3. | +30 | 9 | 9 | | |
| 4. | -30 | 9 | 9 | | |
| 5. | +38 | 16 | 16 | | |
| 6. | -38 | 16 | 16 | | |
| 7. | +40 | 9 | 8 | | |
| 8. | -40 | 9 | 8 | | |

After smoothing:-Now after smoothing the algorithm succeeded in removing one of the fallacies caused due to noise and the other was untreated. The results are as shown below: -

| Table 4 | Nose Localization about Z axis | | | | |
|---|---|---|---|---|---|
| | *Viewpoint around Z axis* | *No. of nose-tips in each angle* | *No. of nose-tips* | *% of success* | *%of failures* |
| 1. | +18 | 13 | 13 | 98.93% | 1.1% |
| 2 | -18 | 13 | 13 | | |
| 3. | +30 | 9 | 9 | | |
| 4. | -30 | 9 | 9 | | |
| 5. | +38 | 16 | 16 | | |
| 6. | -38 | 16 | 16 | | |
| 7. | +40 | 9 | 8 | | |
| 8. | -40 | 9 | 9 | | |

*D. Rotation about X-axis:-*

1) Non-frontal pose:-The fourth test was performed on the 72 face modelvariations selected from the face models of the FRAV3D database. The faces were rotated about x axis (Figure 14). During this test 72 face models (only non -frontal noses rotated about x axis)

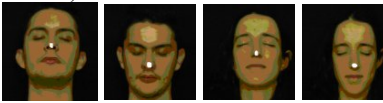

Figure.14. Some samples rotated about x axis from FRAV3D Database

Before Smoothing: - The results obtained before smoothing are shown in the following table:-

| Table 6 | Nose Localization about X axis | | | | |
|---|---|---|---|---|---|
| | *Viewpoint around X axis* | *No. of nose-tips in each angle* | *No. of nose-tips correctly localized* | *% of success* | *%of failures* |
| 1. | +5 | 12 | 12 | 86.11% | 14% |
| 2 | -5 | 12 | 12 | | |
| 3. | +18 | 15 | 15 | | |
| 4. | -18 | 15 | 15 | | |
| 5. | +40 | 9 | 4 | | |
| 6. | -40 | 9 | 4 | | |

After smoothing:- Now after smoothing the algorithm succeeded in removing one of the fallacies caused due to noise and the other was untreated. The results are as:-

| Table 6 | Nose Localization about X axis | | | | |
|---|---|---|---|---|---|
| | *Viewpoint around X axis* | *No. of nose-tips in each angle* | *No. of nose-tips correctly localized* | *% of success* | *%of failures* |
| 1. | +5 | 12 | 12 | 97% | 3% |
| 2 | -5 | 12 | 12 | | |
| 3. | +18 | 15 | 15 | | |
| 4. | -18 | 15 | 15 | | |
| 5. | +40 | 9 | 8 | | |
| 6. | -40 | 9 | 8 | | |

V. COMPARATIVE ANALYSIS

The developed weighted median filtering shows considerable performance over mean filtering because mean filtering removes noise without sharp features. In contrast weighted median filtering provides a good solution even in case of non-uniform meshes and any poses. In this paper, we have presented new methods for triangle mesh denoising:weighted median filtering schemes. Also the proposed weighted mesh median filtering schemes outperforms the conventional mesh smoothing procedures such as the Laplacian smoothing flow and the mean curvature flow. in terms of accuracy and resistance to oversmoothing.

VI. CONCLUSION AND FUTURE SCOPE

The technique which has been presented in this paper on 3D localization has several distinguishing features which would make it suited for face identification in large datasets. In this paper we have practically increased the performance of feature detection algorithm by implementing a smoothing technique. There are a number of other smoothing techniques that are still to be implemented by us. Also our next level of work would be to register the faces i.e. the 3D data models that we are working with, improve the automatic face recognition system by incorporating a more complete model of the human face. It is also being planned to increase the training database to include more profile images which would give us the performance evaluation across large datasets. This should make a more robust template set and increase the systems matching capabilities. With a better model we will also consider methods for matching arbitrary three-dimensional training data thus leading to face registration.